\documentclass[11pt,a4paper]{article}
\usepackage[hyperref]{naaclhlt2019}
\usepackage{times}
\usepackage{latexsym}
\usepackage{graphicx}
\usepackage{subcaption}
\usepackage{multirow}
\usepackage{amsmath}

\usepackage{url}
\usepackage{xcolor}
\usepackage{todonotes} 

\aclfinalcopy 

\title{Reinforcement Learning based Curriculum Optimization for\\Neural Machine Translation}
\author{Gaurav Kumar \\ CLSP, Johns Hopkins \And 
George Foster, Colin Cherry, Maxim Krikun \\ Google AI}

\date{}

\begin{document}
\maketitle
\begin{abstract}
We consider the problem of making efficient use of heterogeneous training data in neural machine translation (NMT). Specifically, given a training dataset with a sentence-level feature such as noise, we seek an optimal {\em curriculum}, or order for presenting examples to the system during training. Our curriculum framework allows examples to appear an arbitrary number of times, and thus generalizes data weighting, filtering, and fine-tuning schemes.  Rather than relying on prior knowledge to design a curriculum, we use reinforcement learning to learn one automatically, jointly with the NMT system, in the course of a single training run. We show that this approach can beat uniform and filtering baselines on Paracrawl and WMT English-to-French datasets by up to +3.4 BLEU, and match the performance of a hand-designed, state-of-the-art curriculum.
  
\end{abstract}

\section{Introduction}

Machine Translation training data is typically heterogeneous: it may vary in
characteristics such as domain, translation quality, and degree of difficulty.
Many approaches have been proposed to cope with heterogeneity, such as
filtering \cite{duh13acl}
or down-weighting  \cite{ D17-1155}
examples that are likely to be noisy or out of domain.
A powerful technique is to control the
curriculum---the order in which examples are presented to the system---as is done
in fine-tuning
\cite{DBLP:journals/corr/FreitagA16},
where training occurs first on general data, and then on 
in-domain data. Curriculum based approaches generalize data filtering
and weighting\footnote{Assuming integer weights.} by allowing examples to be visited multiple times or not at all;
and they additionally potentially 
enable steering the training trajectory toward a
better global optimum than might be attainable with a static
attribute-weighting scheme.

Devising a good curriculum is a challenging task that is typically carried out
manually using prior knowledge of the data and its attributes.
Although powerful heuristics like fine-tuning are helpful, setting hyper-parameters
to specify a curriculum is usually a matter of extensive trial and
error. Automating this process with meta-learning is thus an attractive
proposition. However, it comes with many potential pitfalls such as failing to
match a human-designed curriculum, or significantly increasing training time.

In this paper we present an initial study on meta-learning an NMT curriculum.
Starting from scratch, we attempt to match the performance  of a successful non-trivial reference 
curriculum adopted from
\citet{W18-6314}, in which training gradually focuses on increasingly cleaner
data, as measured by an external scoring function.
Inspired by \citet{Wu18}, we use a reinforcement-learning (RL)
approach involving a learned agent whose task is to
choose a corpus bin, representing a given noise level, at each NMT training step. A
challenging aspect of this task is that choosing only the cleanest bin is
sub-optimal; the reference curriculum uses all the data in the early stages of
training, and only gradually anneals toward the cleanest. Furthermore, we impose
the condition that the agent must learn its curriculum in the course of a single NMT training
run.

We demonstrate that this task is within reach. Our RL agent learns a curriculum
that works as well as the reference, obtaining a similar quality improvement
over a random-curriculum baseline. Interestingly, it does so using a
different strategy from the reference. 
This result opens the door to learning more sophisticated
curricula that exploit multiple data attributes and work with arbitrary
corpora.

\section{Related Work}


Among the very extensive work on handling heterogeneous data in NMT, 
the closest to ours are techniques that re-weight \cite{Chen17} 
or re-order examples to deal with domain mismatch
\cite{vanDerWees17,Sajjad17} or noise \cite{W18-6314}.

The idea of a curriculum was popularized by \citet{Bengio09}, who viewed 
it as a way to improve convergence by presenting heuristically-identified easy
examples first. Two recent papers \cite{Kocmi17,Zhang18} explore similar ideas
for NMT, and verify that this strategy can reduce training time and improve
quality.

Work on meta-learning a curriculum originated with \citet{Tsvetkov16},
who used
Bayesian optimization to learn a linear model for ranking examples in a
word-embedding task. This approach requires a large number of complete training
runs, and is thus impractical for NMT. More recent work has explored bandit
optimization for scheduling tasks in a multi-task problem \cite{Graves17}, and
reinforcement learning for selecting examples in a co-trained classifier
\cite{Wu18}. Finally, \citet{Liu18} apply imitation learning to actively select
monolingual training sentences for labeling in NMT, and show that the learned
strategy can be transferred to a related language pair.


\section{Methods}
\begin{figure}[t]
    \centering
    \includegraphics[width=\linewidth]{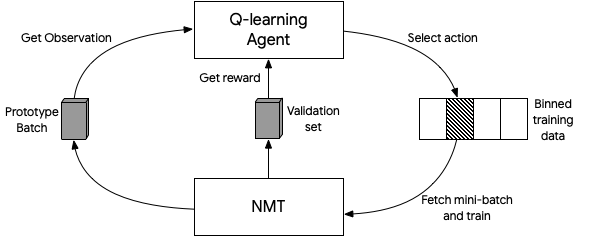}
    \caption{The agent's interface with the NMT system.}
    \label{fig:interface}
\end{figure}

The attribute we choose to learn a curriculum over is noise. 
To determine a per-sentence noise score, we use the {\em contrastive data selection} (CDS) method defined in \citet{W18-6314}.
Given the parameters $\theta_n$ of a model trained on a noisy corpus, 
and parameters $\theta_c$ of the same model fine-tuned on a very small trusted corpus, the score $s(e, f)$ for a translation pair $e, f$ is defined as:
\begin{align}
    s(e, f) = \log p_{\theta_c}(f|e) - \log p_{\theta_n}(f|e)
\end{align}
\citet{W18-6314} show that this correlates very well with human judgments of translation quality.

\subsubsection*{Q-learning for NMT Curricula}
Our agent uses deep Q-learning (DQN) \cite{mnih2015humanlevel}
which is a model-free reinforcement learning procedure.
The agent receives an {\em observation} from the environment and conditions on it to produce an {\em action} which is executed upon the environment. 
It then receives a reward 
representing the {\em goodness} of the executed action.
The agent chooses actions according to a state-action value (Q) function, and
attempts to learn the Q-function so as to maximize expected total rewards.

In our setup, the environment is the NMT system and its training data, as illustrated in Figure~\ref{fig:interface}. We divide the training data into a small number of equal-sized {\em bins} according to CDS scores. At each step, the agent selects a bin 
from which a mini-batch is sampled to train the NMT system.


\subsubsection*{Observation Engineering}
The {\em observation} is meant to be a summary of the state of the environment.
The NMT parameters are too numerous to use as a sensible observation at each time step.
Inspired by \newcite{Wu18}, we propose an observation type which is a function of the NMT system's current performance at various levels of noise.
We first create a {\em prototype batch} by sampling a fixed number of prototypical sentences from each bin of the training data.
At each time step, the observation is the vector containing sentence-level log-likelihoods produced by the NMT system for this prototype batch.

Since the observations are based on likelihood, a metric which aggressively decays at the beginning of NMT training, we use an NMT warmup period to exclude this period from RL training. 
Otherwise, the initial observations would be unlike any that occur later.

\subsubsection*{Reward Engineering} \label{reward}

The {\em reward} is a function of the log-likelihood of the development set of interest.
This implies that the reward naturally decays over time as NMT training proceeds.
A sub-optimal action by the agent may hence receive a larger reward simply by being executed at the beginning of training. 
To combat this, our reward function measures the delta improvement with respect to the average reward received in the recent past.

\begin{figure}[t]
    \centering
    \includegraphics[width=0.8\linewidth]{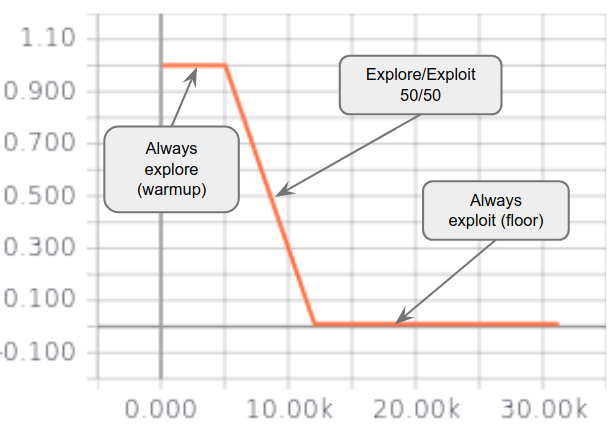}
    \caption{Linearly-decaying {$\epsilon$-greedy} exploration.}
    \label{fig:exploration}
\end{figure}
Our RL agent must balance exploration (choosing an action at random) versus exploitation (choosing the action that maximizes the Q-function). We use a linearly-decaying {$\epsilon$-greedy} exploration strategy (Figure~\ref{fig:exploration}).
This strategy has three phases: (1) The warmup period where we always explore; (2) the decay period where the probability of exploration decreases and exploitation increases; (3) the floor where we always exploit.
Since we do not want to exploit an uninformed Q-function, the duration of exploration needs to be set carefully.
In our experiments, we found that longer decays were useful and the best performance was achieved when the decay was set to about 50\% of the expected NMT training steps.

\section{Experiment Setup} \label{setup}

Our NMT model\footnote{Implemented with Lingvo \cite{Lingvo19}.} 
is similar to RNMT+ \cite{P18-1008}, 
but with four layers in both encoder and decoder. 
Rewards (dev-set likelihood) arrive asynchronously about every 10 training steps.


We use the DQN agent implementation in Dopamine\footnote{\url{github.com/google/dopamine}},
which includes an experience replay buffer to remove temporal correlations
from the observations, among other DQN best practices.
Due to the sparse and asynchronous nature of our rewards, we store observation, action transitions in a temporary buffer until a new reward arrives.
At this point, transitions are moved from the temporary buffer to the DQN agent's replay buffer.
The RL agent is trained after each NMT training step by sampling an RL mini-batch from the replay buffer.
Our RL hyper-parameter settings are listed in the appendix.

Following \citet{W18-6314}, we use the Paracrawl and WMT English-French corpora for our experiments. These contain 290M and 36M training sentences respectively. 
WMT is relatively clean, while a large majority of Paracrawl sentence pairs contain noise. 
We process both corpora 
with BPE, using a vocabulary size of 32k. 
The WMT newstest 2010-2011 corpus is used as trusted data for CDS scores, which are computed using the models and procedure described in \citet{W18-6314}. Both corpora are split into 6 equal-sized bins according to CDS score. For the {\em prototype batch} used to generate observations, we extracted the 32 sentences whose CDS scores are closest to the mean in each bin, giving a total of 192 sentences.
We use WMT 2012-2013 for development and WMT 2014 for test, and report tokenized, 
naturally-cased BLEU scores from the test checkpoint closest to the highest-BLEU dev checkpoint.
Since reinforcement learning methods can be unstable, all models were run twice with different random seeds, and the model with the best score on the dev set was chosen.

\begin{table}[h]
    \centering
    \begin{tabular}{|c|c|c|} \hline
         & \bf{Paracrawl} & \bf{WMT} \\ \hline
         \multicolumn{3}{|c|}{\bf{Uniform baselines}} \\ \hline
         Uniform & 34.1 & 37.1 \\ \hline
         Uniform (6-bins) & 34.8 & - \\ \hline 
         Uniform (bookends) & 35.0 & 34.8 \\ \hline\hline
         \multicolumn{3}{|c|}{\bf{Heuristic baselines}} \\ \hline
         Filtered (33\%/20\%) & 37.0 & 38.3 \\ \hline
         Fixed $\epsilon$-schedule & 36.9 & 37.7 \\ \hline
         Online & 37.5 & 37.7 \\ \hline\hline
         \multicolumn{3}{|c|}{\bf{Learned curricula}} \\ \hline
         Q-learning (bookends) & 36.8 & 36.3 \\ \hline
         Q-learning (6-bins) & 37.5 & 38.4 \\ \hline
    \end{tabular}
    \caption{BLEU scores on Paracrawl and WMT En-Fr datasets with uniform, heuristic and learned curricula.}
    \label{tab:results}
\end{table}

\section{Results}
Our results are presented in Table~\ref{tab:results}. {\bf Uniform baselines} consist of: {\em Uniform} -- standard NMT training; {\em Uniform (6-bins)} -- uniform sampling over all bins; and {\em Uniform (bookends)} -- uniform sampling over just the best and worst bins. Surprisingly, 6-bins performs better than the standard NMT baseline. We hypothesize that this can be attributed to more homogeneous mini-batches.


{\bf Heuristic baselines} are: 
{\em Filtered} -- train only on the highest-quality data as determined by CDS scores: top 20\% of the data for Paracrawl, top 33\% for WMT. {\em Fixed $\epsilon$-schedule}  -- we use the $\epsilon$-decay strategy of our best RL experiment but always choose the cleanest bin when we exploit. 
{\em Online} -- the online schedule from \citet{W18-6314} adapted to the 6-bin setting. We verified experimentally that our performance matched the original schedule, which did not use hard binning.


{\bf Learned curricula} were trained over 2 bookend bins and all 6 bins. On the Paracrawl dataset, in the 2-bin setting, the learned curriculum beats all uniform baselines 
and almost matches the optimized filtering baseline.
With 6-bins, it beats all uniform baselines by up to 2.5 BLEU and matches the hand-designed online baseline of \citet{W18-6314}.
On WMT, with 2 bins, the learned curriculum beats the 2-bin baseline, but not the uniform baseline over all data. With 6 bins, the learned curriculum beats the uniform baseline by 
1.5 BLEU, and matches the filtered baseline, which in this case outperforms the online curriculum by 0.6 BLEU.

Our exploration strategy for Q-learning (see Figure~\ref{fig:exploration}) forces the agent to visit all bins during initial training, and only gradually rely on its learned policy. This mimics the gradual annealing of the online curriculum, so one possibility is that the agent is simply choosing the cleanest bin whenever it can, and its good performance comes from the enforced period of exploration. However, the fact that the agent beats the fixed $\epsilon$-schedule (see Table~\ref{tab:results}) described above on both corpora makes this unlikely.


\section{Analysis}

\begin{table}[]
    \centering
    \begin{tabular}{|c|c|c|} \hline
         Reward/Observations & Default & Fixed \\ \hline
         Default & \bf{37.5} & 36.2\\ \hline
         Fixed & 32.5 & - \\ \hline
    \end{tabular}
    \caption{BLEU scores on ablation experiments with fixed rewards or observations.}
    \label{tab:information}
\end{table}
Task-specific reward and observation engineering is critical when building an RL model. We performed ablation experiments to determine if the rewards and observations we have chosen contain information which aids us in the curriculum learning task. Table~\ref{tab:information} shows the results of our experiments. The fixed reward experiments were conducted by replacing the default delta-perplexity based reward with a static reward which returns a reward of one when the cleanest bin was selected and zero otherwise. The fixed observation experiments used a static vector of ones as input at all time steps. Both settings perform worse than our default setup, indicating that both the chosen observation and reward provide information to help with this task.

\subsection{What did the agent learn?}
\begin{figure}[ht!]
    \centering
    \begin{subfigure}[b]{0.21\textwidth}
        \includegraphics[width=\textwidth]{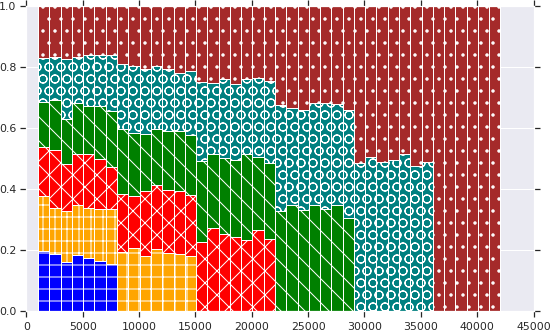}
        \caption{Online}
    \end{subfigure}
    \begin{subfigure}[b]{0.21\textwidth}
        \includegraphics[width=\textwidth]{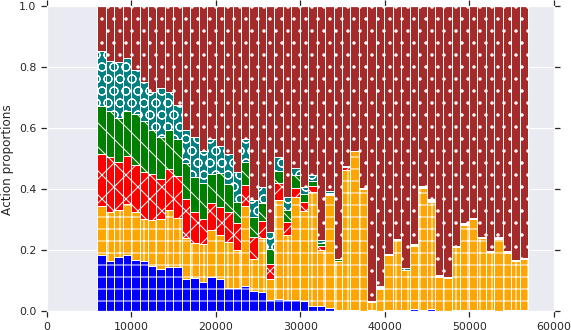}
        \caption{RL learned}
    \end{subfigure}
    \caption{Online policy from \citet{W18-6314} compared to the RL policy. Lower bins contain more noise.}
    \label{fig:policy}
\end{figure}
Figure~\ref{fig:policy} shows a coarse visualization of the hand-optimized policy of \citet{W18-6314}, adapted to our 6-bin scenario, compared to the Q-learning policy on the same scenario. The former, by design, telescopes towards the clean bins. Note that the latter policy is masked by the agent's exploration schedule, which slowly anneals toward nearly complete policy control, beginning at step 30,000.
After this point, the learned policy takes over and continues to evolve.
This learned policy has little in common with the hand-designed one. Instead of focusing on a mixture of the clean bins, it focuses on the cleanest bin and the second-to-noisiest. We hypothesize that returning to the noisy bin acts as a form of regularization, though this requires further study. 



\section{Conclusion}
We have presented a method to learn the curriculum for presenting training samples to an NMT system.
Using reinforcement learning, our approach learns the curriculum jointly with the NMT system during the course of a single NMT training run.
Empirical analysis on the Paracrawl and WMT English-French corpora shows that this approach beats the uniform sampling and filtering baseline by large margins. In addition, we were able to match a state-of-the-art hand designed curriculum on Paracrawl and beat it on WMT.

We see this a first step toward enabling NMT systems to manage their own training data.
In the future, we intend to improve our approach by eliminating the static exploration schedule and binning strategy, and extend it to handle additional data attributes such as domain, style and grammatical complexity.

\clearpage

\bibliography{naaclhlt2019}
\bibliographystyle{acl_natbib}

\clearpage

\appendix
\section{Appendix}
\label{sec:appendix}
\begin{figure*}[]
    \centering
    \begin{subfigure}[b]{0.45\textwidth}
        \includegraphics[width=\textwidth]{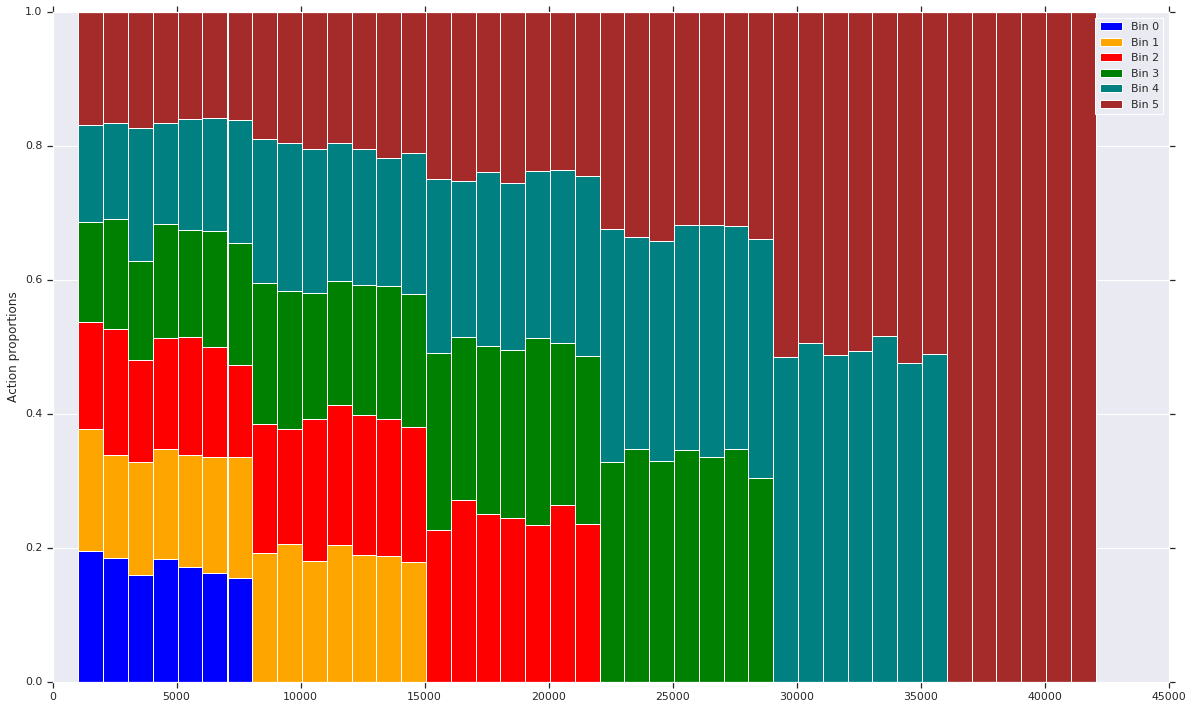}
        \caption{Telescoping}
    \end{subfigure}
    \begin{subfigure}[b]{0.45\textwidth}
        \includegraphics[width=\textwidth]{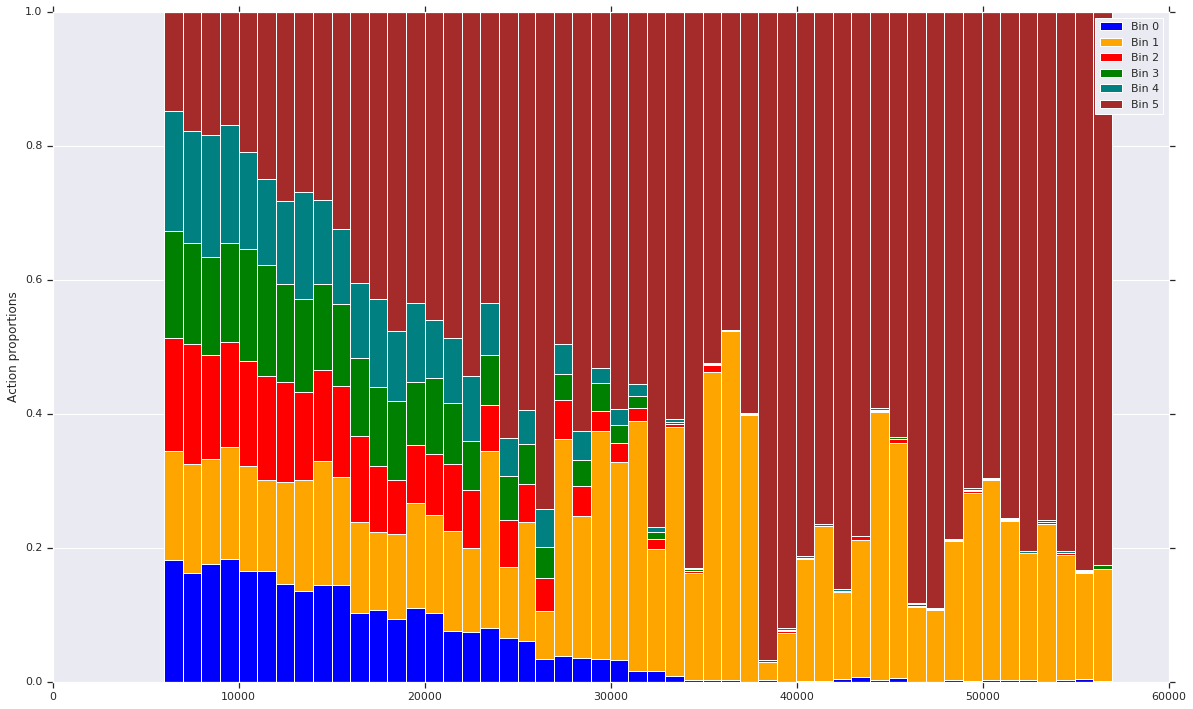}
        \caption{RL learned}
    \end{subfigure}
    \caption{Policies learned by the RL agent on the Paracrawl En-Fr corpus compared against the telescoping policy from \citet{W18-6314}. Lower bins contain more noise.}
    \label{fig:policies-paracrawl}
\end{figure*}

\begin{figure*}[]
    \centering
    \begin{subfigure}[b]{0.45\textwidth}
        \includegraphics[width=\textwidth]{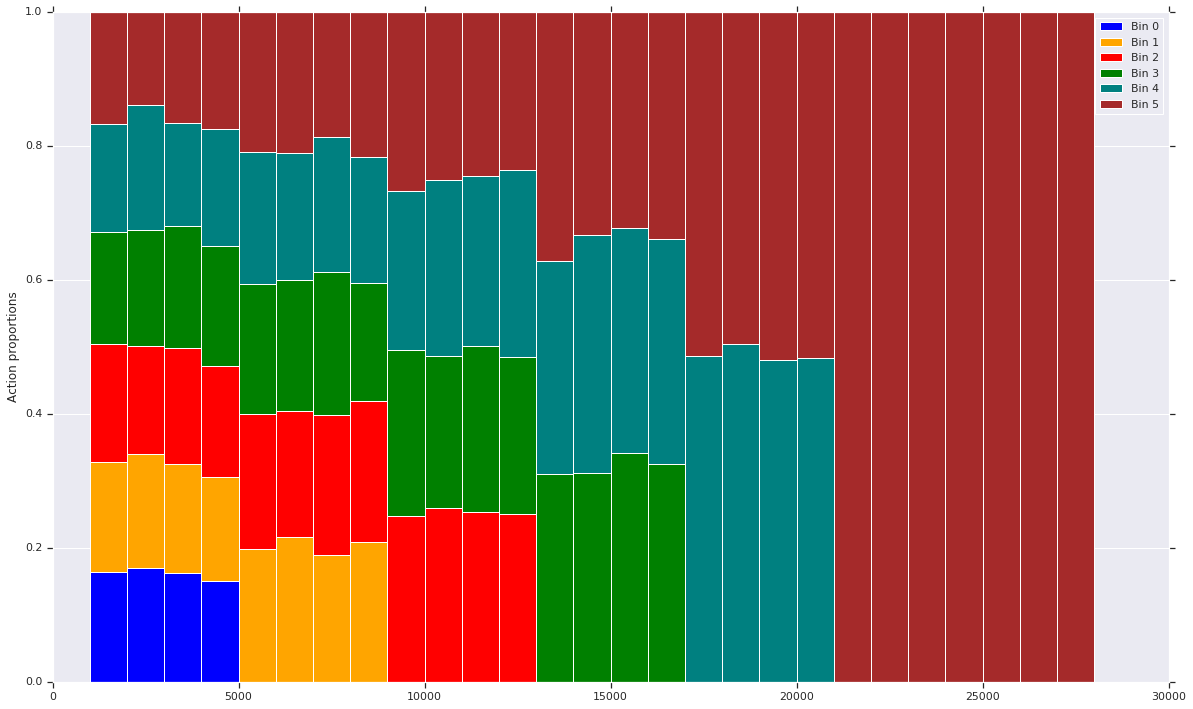}
        \caption{Telescoping}
    \end{subfigure}
    \begin{subfigure}[b]{0.45\textwidth}
        \includegraphics[width=\textwidth]{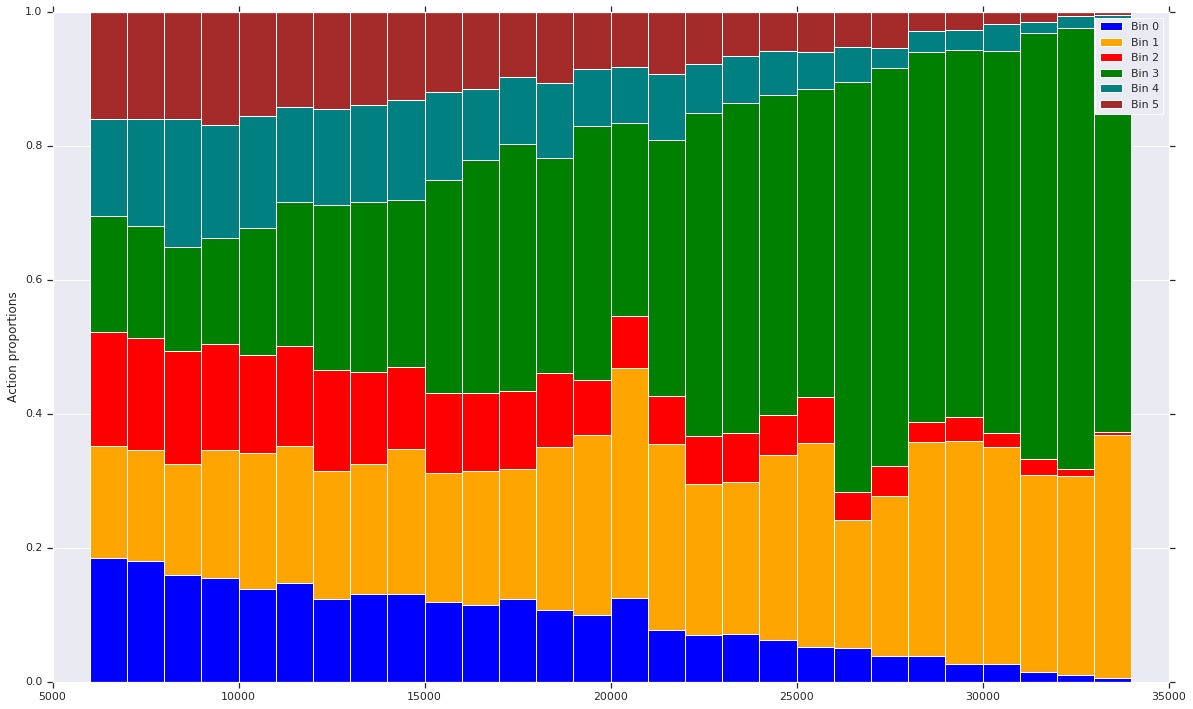}
        \caption{RL learned}
    \end{subfigure}
    \caption{Policies learned by the RL agent on the WMT En-Fr corpus compared against the telescoping policy from \citet{W18-6314}. Lower bins contain more noise.}
    \label{fig:policies-wmt}
\end{figure*}

\begin{figure*}[]
    \centering
    \begin{subfigure}[b]{0.45\textwidth}
        \includegraphics[width=\textwidth]{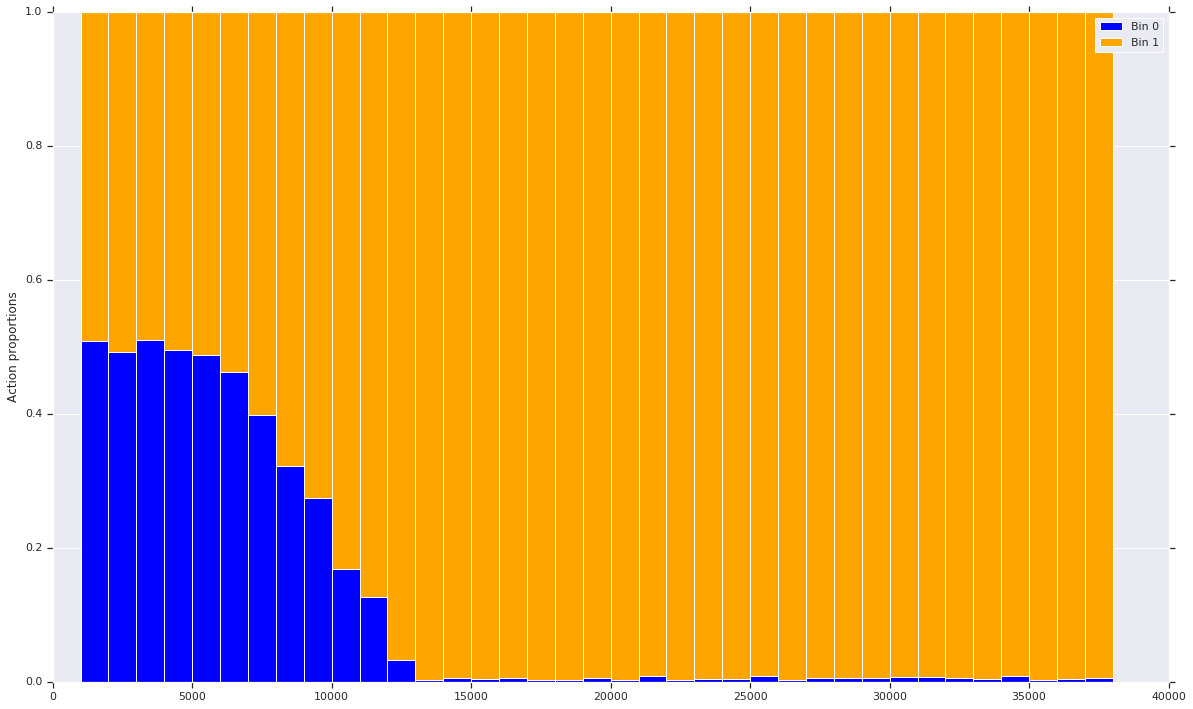}
        \caption{RL Learned (Paracrawl)}
    \end{subfigure}
    \begin{subfigure}[b]{0.45\textwidth}
        \includegraphics[width=\textwidth]{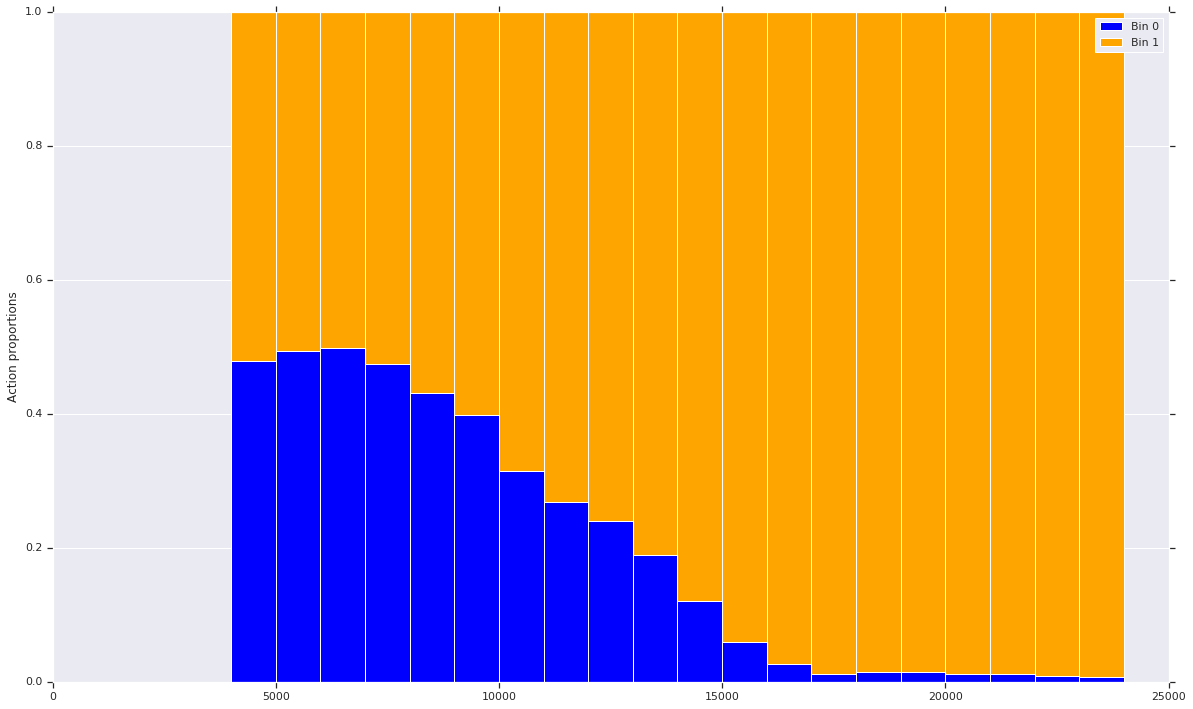}
        \caption{RL learned (WMT)}
    \end{subfigure}
    \caption{Policies learned by the RL agent on the 2-bin task on the Paracrawl and WMT En-Fr datasets. Lower bins contain more noise.}
    \label{fig:policies-2bin}
\end{figure*}

\subsection{Q-learning hyper-parameters}
\begin{itemize}
    \item Observations: We sample 32 prototype sentences from each bin to create a {\em prototype batch} of 192 sentences.
    \item Q-networks: The two Q-networks were MLPs with 2 x 512-d hidden layers each. A ${\tanh}$ activation function was used.
    \item RL optimizer: We used RMSProp with a learning rate of 0.00025 and a decay of 0.95 and no momentum.
    \item NMT warmup : 5000 steps (no transitions from this period are recorded).
    \item Stack size: We do not stack our observations for the RL agent (i.e., stack size = 1).
    \item Exploration strategy : We use a linearly decaying epsilon function with decay period set to 25k steps. The decay floor was set to 0.01.
    \item Discount gamma : 0.99
    \item Update horizon : 2
    \item Minimum number of transitions in replay buffer before training starts: 3000
    \item Update period (how often the online Q-network is trained): 4 steps
    \item Target update period (how often the target Q-network is trained): 100 steps
    \item The window for the delta-perplexity reward was 1.
\end{itemize}

\subsection{Learned Policies}
Figures~\ref{fig:policies-paracrawl}, ~\ref{fig:policies-wmt} and ~\ref{fig:policies-2bin} show coarse representations of the policies learned by the Q-learning agent on the Paracrawl and WMT English-French datasets. Each column in the figures represents the relative proportion of actions taken (bins selected) averaged over a thousand steps and the actions go from noisy to clean on the y-axis. Each policy starts from a uniform distribution over actions. Some salient aspects of the learned policies are listed below.

\begin{enumerate}
    \item All learned curricula differ significantly from the hand-designed policies.
    \item The RL curriculum learned for Paracrawl (Figure~\ref{fig:policies-paracrawl}) focus on two bins during exploitation (choose action using the trained Q-function). Surprisingly, these are not the two cleanest bins but a mixture of the cleanest and the second-to-noisiest bin.
    \item The RL curriculum learned for WMT (Figure~\ref{fig:policies-paracrawl}) is closer to a uniform distribution over actions for a long duration. This makes sense since the data from WMT is mostly homogeneous with respect to noise. When the agent does decide to exploit some bins more often, they are not the cleanest ones, but the 1st and 4th bin instead.
    \item Figure~\ref{fig:policies-2bin} shows the policies learned on the bookend task for Paracrawl and WMT; the only two bins available contain the noisiest and cleanest portion of the corpus. The RL agent very quickly learns that there is an optimal bin to choose in this task and converges to consistently exploiting it. We consider this a sanity check of curriculum learning methods.
\end{enumerate}

\end{document}